\documentclass[runningheads]{llncs}
\pdfoutput=1
\usepackage{graphicx}
\usepackage{amsmath}
\usepackage{amssymb}
\usepackage{multirow}
\usepackage{caption}
\usepackage{subfigure}
\usepackage{color}
\usepackage{url}  %Required
\usepackage{booktabs}
\usepackage{cancel}

\newcommand\blfootnote[1]{% 
\begingroup 
\renewcommand\thefootnote{}\footnote{#1}% 
\addtocounter{footnote}{-1}% 
\endgroup 
}

\begin{document}
\title{VGCN-BERT: Augmenting BERT with  Graph Embedding for Text Classification
%\thanks{Supported by organization x.}
}
\titlerunning{VGCN-BERT}

\author{Zhibin Lu \and Pan Du \and Jian-Yun Nie}
%\author{Zhibin Lu\orcidID{0000-0003-3133-5770} \and Pan Du \and Jian-Yun Nie}
% ORCID of Zhibin Lu: http://orcid.org/0000-0003-3133-5770
%
\authorrunning{Lu et al.}
\institute{
University of Montreal, Montreal, Canada \\
\email{\{zhibin.lu, pan.du\}@umontreal.ca, nie@iro.umontreal.ca} }
\maketitle              % typeset the header of the contribution
\begin{abstract}
Much progress has been made recently on text classification with methods based on neural networks. In particular, models using attention mechanism such as BERT have shown to have the capability of capturing the contextual information within a sentence or document. However, their ability of capturing the global information about the vocabulary of a language is more limited. This latter is the strength of Graph Convolutional Networks (GCN). 
In this paper, we propose VGCN-BERT model which combines the capability of BERT with a Vocabulary Graph Convolutional Network (VGCN). Local information and global information interact through different layers of BERT, allowing them to influence mutually and to build together a final representation for classification.
%and Self-Attention Encoder (BERT). 
In our experiments on several text classification datasets, our approach outperforms BERT and GCN alone, and achieve higher effectiveness than that reported in previous studies.

\keywords{Text classification \and BERT \and Graph Convolutional Networks.}
\end{abstract}
%
%
%
%%%%%%%%%%%%%%%%%%%%%%%%%%%%%%%%%
%%        Introduction         %%
%%%%%%%%%%%%%%%%%%%%%%%%%%%%%%%%%
%
%
\section{Introduction}
\blfootnote{\copyright{} Springer Nature Switzerland AG 2020

J. M. Jose et al. (Eds.): ECIR 2020, LNCS 12035, pp.369-382, 2020. 

https://doi.org/10.1007/978-3-030-45439-5}

Text classification is a fundamental problem in natural language processing (NLP) and has been extensively studied in many real applications. In recent years, we witnessed the emergence of text classification models based on neural networks such as  convolutional neural networks (CNN)~\cite{kim2014convolutional}, recurrent neural networks (RNN)~\cite{hochreiter1997long}, and various models based on attention~\cite{vaswaniAttention}. BERT~\cite{devlin2018BERT} is one of the self-attention models that uses multi-task pre-training technique based on large corpora. It often achieves excellent performance, compared to CNN/RNN models and traditional models, in many tasks~\cite{devlin2018BERT} such as Named-entity Recognition (NER), text classification and reading comprehension. 

The deep learning models excel %traditional models 
by embedding both semantic and syntactic information in a learned  representation. However, most of them are known to be limited in encoding long-range dependency  information of the text~\cite{battaglia2018relational}. The utilization of self-attention helps alleviate this problem, but the problem still remains. The problem stems from the fact that the representation is generated from a sentence or a document only, without taking into account explicitly the knowledge about the language (vocabulary). 
%hence for sentences with ambiguous expressions or inconspicuous words, the new representation may be insufficient for the classifier to make a judgement. 
For example, in the movie review below:

\textit{``Although it's a bit smug and repetitive, this documentary engages your brain in  \underline{a way few current films do}.''}

Both negative and positive opinions appear in the sentence. Yet the positive attitude \sloppy \textit{\underline{``a way few current films do''}} expresses a very strong opinion of the innovative nature of the movie in an implicit way. Without connecting this expression more explicitly to the meaning of \textit{``innovation"} in the context of movie review comments, the classifier may underweight this strong opinion and the sentence may be wrongly classified to be negative. On this example, self-attention that connects the expression to other tokens in the sentence may not help.

In recent studies,  approaches have also been developed to take into account the global information between words and concepts. The most representative work is Graph Convolutional Networks (GCN)~\cite{kipf2017semi} and its variant Text GCN~\cite{yao2019graph}, in which words in a language are connected in a graph. By performing convolution operations on neighbor nodes in the graph, the representation of a word will incorporate those of the neighbors, allowing to integrate the global context of a domain-specific language to some extent. For example, the meaning of \textit{``new"} can be related to that of \textit{``innovation''} and \textit{``surprised''} through the connections between them.
%in the context, so that the meaning of ``\textit{\underline{a way few current films do}}'' in this context is associated with vocabulary (and meaning) of \textit{``innovation''} and \textit{``supprised''}, making it a much easier case for classification. 
However, GCNs that only take into account the global vocabulary information may fail to capture local information (such as word order), which is very important in understanding the meaning of a sentence. This is shown in the following examples, where the position of \textit{``work"} in the sentence will change the meaning depending on its context:
%, the local information, such as the order between words, is missing in the convolution. The order between words is also important in text information since the meaning can be very different, even reversed, if the order of the words is changed, take another two review comments for example:
\begin{itemize}
    \item \textit{``skip \underline{work} to see it at the first opportunity.''}
    \item \textit{``skip to see it, \underline{work} at the first opportunity.''}
\end{itemize}

%In this paper, inspired by Text GCN~\cite{yao2019graph} and self-attention mechanism in BERT, 
In this paper, inspired by GCN~\cite{kipf2017semi,yao2019graph} and self-attention mechanism in BERT, we propose to combine the strengths of both mechanisms in the same model. We
first construct a graph convolutional network on the vocabulary graph based on the word co-occurrence information, which aims at encoding the global information of the language, then feed the graph embedding and word embedding together to a self-attention encoder  in  BERT. The word embedding and graph embedding then interact with each other through the self-attention mechanism while learning the classifier. This way, the classifier can not only make use of both local information and global information, but also allow them to guide each other via the attention mechanism so that the final representation built up for classification will integrate gradually both local and global information. We also expect that the connections between words in the initial vocabulary graph can be spread to more complex expressions in the sentence through the layers of self-attention.

We call the proposed model VGCN-BERT. 
Our source code is available at~\url{https://github.com/Louis-udm/VGCN-BERT}.

We carry out experiments on 5 datasets of different text classification tasks (sentiment analysis, grammaticality detection, and hate speech detection). On all these datasets, our approach is shown to outperform  BERT and GCN alone. 

The contribution of this work is twofold:
%VGCN-BERT is enabled and demonstrated by the following technical contributions:
\begin{itemize}
    \item \textit{Combining global and local information}: There has not been much work trying to combine local information captured by BERT and global information of a language. We demonstrate that their combination is beneficial.
    \item \textit{Interaction between local and global information through attention mechanism}: We propose a tight integration of local information and global information, allowing them to interact through different layers of networks.
\end{itemize}

%%%%%%%%%%%%%%%%%%%%%%%%%%%%%%%%%
%% Background and Related Work %%
%%%%%%%%%%%%%%%%%%%%%%%%%%%%%%%%%

\section{Related Work}

\subsection{Self-Attention and BERT}

As aforementioned, attention mechanisms~\cite{yang2016hierarchical,wang2016attention} based on various deep neural networks, in particular the self-attention mechanism proposed by Vaswan et al.~\cite{vaswaniAttention}, have greatly improved the performance in text classification tasks.
The representation of a word acquired through self-attention can incorporate the relationship between the word and all other words in a sentence by fusing the representations of the latter.

BERT (Bidirectional Encoder Representations from Transformers) \cite{devlin2018BERT}, which leverages a multi-layer multi-head self-attention (called transformer) together with a positional word embedding, is one of the most successful deep neural network model for text classification in the past years. The attention %(a layer-superimposed self-attention) 
mechanism in each layer of the encoder enhances the new representation of the input data with contextual information by paying multi-head attentions to different parts of the text. A pre-trained BERT model based on 800M words from BooksCorpus and 2,500M words from English Wikipedia is made available. It has also been widely used in  many NLP tasks, and has proven effective. However, as most of other attention-based deep neural networks, BERT mainly focuses on local consecutive word sequences, which provides local context information. That is, a word is placed in its context, and this generates a contextualized representation. However, it may be difficult for BERT to account for the global information of a language. 

\subsection{Graph Convolutional Networks (GCN)}

Global relations between words in a language can be represented as a graph, in which words are nodes and edges are relations. Graph Neural Network (GNN)~\cite{cai2018comprehensive,battaglia2018relational} based on such a graph is good at capturing the general knowledge about the words in a language. A number of variants of GNN have been proposed and applied to text classification tasks~\cite{henaff2015deep,defferrard2016convolutional,kipf2017semi,peng2018large,DBLPconfaclZhangLS18}, of which Kipf et al.\cite{kipf2017semi} creatively presented Graph Convolutional networks (GCN) based on spectral graph theory. GCN first builds a symmetric adjacency matrix based on a given relationship graph (such as a paper citation relationship), and then the representation of each node is fused according to the neighbors and corresponding relationships in the graph during the convolution operation. 
 
Text GCN is a special case of GCN for text classification  proposed by Yao et al.~\cite{yao2019graph} recently. Different from general GCN, it is based on a heterogeneous graph where both words and documents are nodes. The relationships among nodes, however, are measured in three different ways, which are co-occurrence relations among words, tf-idf measure between documents and words, and self similarity among documents.
In terms of convolution, Text GCN uses the same algorithm as GCN.
%, but does not feed the document features during training, because the document features are the nodes integrated into the graph.
GCN and its variants are good at convolving the global information in the graph into a sentence, but they do not take into account local information such as the order between words. When word order and other local information are important, %as in many NLP problems, 
GCN may be insufficient. Therefore, it is natural to combine GCN with a model capturing local information such as BERT.
%in its output representation. 

\subsection{Existing Combinations of GCN and BERT}

Some recent studies have combined GCN with BERT. Shang et al.~\cite{vanillaGCNtoBERT} applied a combination to the medication recommendation task, which predict a medical code given the electronic health records (EHR), i.e., a sequence of historical medical codes, of a patient. They first embed the medical codes from a medical ontology using Graph Attention Networks (GAT), then feed the embedding sequence of the medical code in an EHR into BERT for code prediction. Nevertheless, the order in the code sequence is discarded in the transformer since it is not applicable in their scenario, making it incapable of capturing all the local information as in our text classification tasks. 

Jong et al.~\cite{vanillaBertGCN} proposed another combination to the citation recommendation task using paper citation graphs. This model simply concatenates the output of GCN and the output of BERT for downstream predictive tasks. We believe that  interactions between the local and global information are important and can benefit the downstream prediction tasks. In fact, through layers of interactions, one could allow the information captured in GCN be applied to the input text, and the representation of the input text be spread over GCN. This will produce the effect we illustrated in the earlier example of movie review (\emph{a way few current films do} vs. \emph{innovation}).
%The mutual influence between local (input text) and global (general knowledge about words) information may lead to a richer representation, than that obtained from the two parts separately. 
This is the approach we propose in this paper.
%by mutually guiding each other to pay different attentions to different parts accordingly, as is demonstrated in the previous example.

One may question about the necessity to explicitly use graph embedding to cope with global dependency information, as some studies~\cite{DBLP:conf/nips/LevyG14,iclr2020mfemb} have shown that word embedding trained on a corpus, such as Word2Vec~\cite{DBLP:journals/corr/abs-1301-3781}, GloVe~\cite{glove}, FastText~\cite{DBLP:conf/eacl/GraveMJB17}, can capture some global connections between words in a language. We believe that a vocabulary graph can still provide additional information given the fact that the connections between words observed in word embeddings are limited within a small text window (usually 5 words). Long-range connections are missing.
In addition, by building a vocabulary graph on an application-specific document collection, one can capture application-dependent dependencies, in addition to the general dependencies in the pre-trained models.
%the application can provide new information. 
%, which is trained by using unsupervised learning methods with large-scale corpus, reflects more generally co-occurrence information, rather than task-specific. Nevertheless, when fine-tuning the model using a small-scale dataset for a specific task, it is not easy to fine-tune the word embedding containing the co-occurrence information of specific tasks through the skip-gram model. On the contrary, we usually fine-tune the weights of the latter part of the model for specific tasks by supervised learning.
%Our vocabulary graph aims to solve these problems by constructing a vocabulary graph based on co-occurrences of words within a larger context (i.e. sentence), and on an application-specific document collection.
%computing their PMI based on co-occurrences within documents, and on a document collection.
%through the VGCN operation, directly allows the document-level co-occurrence information from specific task to participate in the training of the fine-tuning stage.

%%%%%%%%%%%%%%%%%%%%%%%%%%%%%%%%%
%%           Method            %%
%%%%%%%%%%%%%%%%%%%%%%%%%%%%%%%%%

\section{Proposed Method}

\begin{figure*}[!htp]
  \centering
  \includegraphics[height=27 mm]{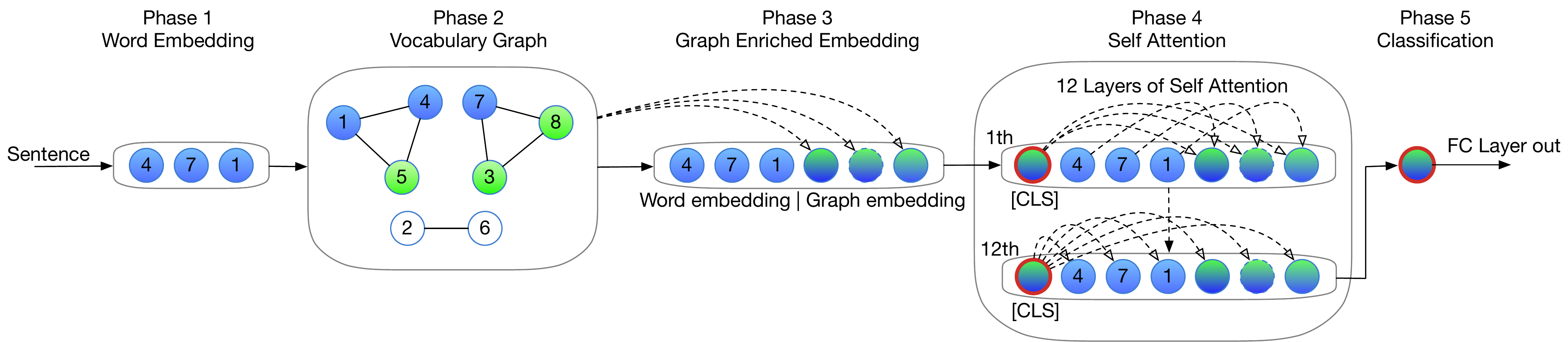}
  \caption{Illustration of VGCN-BERT. The embeddings of input sentence (Phase 1) are combined with the vocabulary graph (Phase 2) to produce a graph embedding, which is concatenated to the input sentence (Phase 3). Note that from the vocabulary graph, only the part relevant to the input is extracted and embedded. In Phase 4, several layers of self-attention are applied to the concatenated representation, allowing interactions between word embeddings and graph embedding. The final embedding at the last layer is fed in a fully connected layer (Phase 5) for classification. 
%  Assume a vocabulary graph which contains seven words and their relationship of NPMI is shown in the rectangle box, assume an input sentence it's vocabulary ID sequence is (4,7,1). Blue represents the currently entered words, green represents the associated words in graph, and blue-green blend represents the convolutional representation. In phase2 and phase3, the Vocabulary GCN module will involve the vocabulary graph to the word embedding of (4,7,1).%, in this case (3,5,8) will be convoluted in but (2,6) will not.
%  Then the convoluted output from VGCN module will be reduced to several (the hyperparameter=3 in this demonstration) embedding vector which has the same shape as word embedding. After enrich by the global graph embedding, in the 12-layer self-attention encoder, the word embedding will be fully cross-fused with the graph embedding, and in the last layer, the [CLS] embedding will pay different attention to all other embeddings as well as graph embeddings, and finally output a sentence embedding with global information for classification.
}
  \label{fig:framework}
\end{figure*}

%Our goal is to obtain global language information for enhancement while retaining the original local information. 
The global language information can take multiple forms. In this paper, we consider lexical relations in a language, i.e. a vocabulary graph. Any vocabulary graph can be used to complement BERT (e.g. Wordnet). In this paper, we consider a graph constructed using word co-occurrences with documents. Local information from a text is captured by BERT. The interaction between them is achieved by first selecting the relevant part of the global vocabulary graph according to the input sentence and transforming it into an embedding representation. We use multiple layers of attention mechanism on concatenated representation of input text and the graph. These processes are illustrated in 
Fig. \ref{fig:framework}.
We will provide more details in the following subsections.

\subsection{Vocabulary Graph}

Our vocabulary graph is constructed using normalized point-wise mutual information (NPMI)~\cite{npmi}, as shown in Equation \ref{eq:npmi}:

\begin{equation}\label{eq:npmi}
\text{NPMI}(i, j) = -\frac{1}{\log p(i,j)} \log \frac{p(i, j)}{p(i)p(j)}
\end{equation}

\noindent where $i$ and $j$ are words, $p(i, j) = \frac{\#W(i, j)}{\#W}$, $p(i) = \frac{\#W(i)}{\#W}$, $\#W(*)$ is the number of sliding windows containing a word or a pair of words,
%in a corpus that contain word $i$, $\#W(i, j)$ is the number of sliding windows that contain both word $i$ and $j$, 
and $\#W$ is the total number of sliding windows. To obtain  long-range dependency, we set the window to the whole sentence. The range of value of NPMI is [-1,1]. A positive NPMI value implies a high semantic correlation between words, while a negative NPMI value indicates little or no semantic correlation. In our approach, we create an edge between two words if their NPMI is larger than a threshold. Our experiments show that the performance is better when the threshold is between $0.0$ and $0.3$.

% The advantage of calculating NPMI is convenient for threshold settings. Since the range of NPMI is [-1,1], we can use the threshold as a hyperparameter to set the best value for the model. in addition, when the corpus is large, in order to maintain the sparsity of the matrix of $A$, we can set a threshold of NPMI to take a part of the [-1,1] interval. In our case, we set the threshold to 0.2. 

\subsection{Vocabulary GCN}
A general GCN~\cite{kipf2017semi} is a multi-layer (usually 2 layers) neural network that convolves directly on a graph and induces embedding vectors of nodes based on properties of their neighborhoods. Formally, consider a graph $G = (P, E)$ \footnote{In order to distinguish from notations $(v,V,|V|)$ of vocabulary, this paper uses notations $(p,P,|P|)$ to represent the point(vertex) of the graph. }, where $P$ (with $|P| = n$) and $E$ are sets of nodes and edges, respectively. 
% In GCN, the graph is a self-loop graph, where every node is assumed to be connected to itself, i.e., $(p, p) \in E$ for any $p$.
%Usually, people use the adjacency matrix $A$ and its degree matrix $D$ to represent graph $G$, where diagonal elements of $A$ are set to $1$ because of self-loops, and $D$ is a diagonal matrix, $D_{ii} = \sum_j A_{ij}$. 
For a single convolutional layer of GCN, %with a single convolutional layer, 
the new representation is calculated as follows:
\begin{equation}
  H = \tilde{A}XW,
  \label{eq1_gcn_1layer}
\end{equation}

\noindent where $X \in \mathbb{R}^{n \times m}$ is the input matrix with $n$ nodes and $m$ dimensions of the feature, $W \in \mathbb{R}^{m \times h}$ is a weight matrix, $\tilde{A} = D^{-\frac{1}{2}}A D^{-\frac{1}{2}}$ is the normalized symmetric  adjacency matrix, where $D_{ii} = \sum_j A_{ij}$. The normalization operation for $A$ is to avoid numerical instabilities and exploding/vanishing gradients when used in a deep neural network model~\cite{kipf2017semi}.

% The graph nodes of GCN are ``task entities'' (such as documents that need to be classified), therefore, GCN is a transductive learning mode, that is, all entities including test set, is convolved by the whole graph in one time, so that the neighbors representations of each entity can be aggregated. But in our case, a document is no longer a node of the graph, it is seen as consisting of words, and the graph has changed from document nodes to word nodes (vocabulary graph), then our convolution operator has changed from all documents one time to all words of one document one time. Because our purpose is to convolve the neighbor words in the document, not the neighor documents in the corpus. Thus, for a single document, assuming the document is a row vector consists of words in vocabulary, then we derive the formula~\ref{eq1_gcn_1layer},

The graph nodes of GCN are ``task entities'' such as documents that need to be classified. It requires all entities, including those from training set, validation set, and test set, to be presented in the graph, so that no node representation is missing in downstream tasks. This limits the application of GCN in many predictive tasks, where the test data is unseen during the training process.

% But in our case, a document is no longer a node of the graph, it is seen as consisting of words, and the graph has changed from document nodes to word nodes (vocabulary graph), then our convolution operator has changed from all documents one time to all words of one document one time. Because our purpose is to convolve the neighbor words in the document, not the neighor documents in the corpus. Thus, for a single document, assuming the document is a row vector consists of words in vocabulary, then we derive the formula~\ref{eq1_gcn_1layer},

In our case, we aim to convolve the related words instead of the documents in the corpus for classification. Therefore, the graph of our proposed GCN  is constructed on the vocabulary instead of the documents. Thus, for a single document, assuming the document is a row vector $\vec{x}$ consisting of words in the vocabulary, a layer of convolution is defined in equation~\ref{eq_vgcn_1}:

\begin{equation}
  \vec{h} = (\tilde{A}\vec{x}^T)^TW=\vec{x}\tilde{A}W,
  \label{eq_vgcn_1}
\end{equation}

\noindent where $\tilde{A}^T=\tilde{A}$ represent the vocabulary graph. $\vec{x}\tilde{A}$ extracts the part of vocabulary graph relevant to the input sentence $\vec{x}$. %Note that every column of $X$ in Equation~\ref{eq1_gcn_1layer} is a corpus size vector, i.e. a column vector of all documents, but  $\vec{x}^T$ here is a vocabulary size column vector corresponding to a single document. 
$W$ holds the weights of the hidden state vector for the single document, with dimension $|V|\times h$. Then for $m$ documents in a mini-batch, the one-layer  graph convolution in Equation~\ref{eq_vgcn_1} becomes:

\begin{equation}
  H = X\tilde{A}W,
  \label{eq_vgcn_2}
\end{equation}

% It is a one-layer vocabulary graph convolution for the formula~\ref{eq_vgcn_2}. In order to obtain the non-linear fitting ability of the neural network, we add the non-linear activation function ReLU and another fully connected layer to the formula~\ref{eq_vgcn_2}. Therefore, the vocabulary graph convolutional formula is as follows:
\noindent and the corresponding 2-layer Vocabulary GCN with ReLU  function is as follows:
\begin{equation}
    \textbf{VGCN} = \text{ReLU}( X_{mv} \tilde{A}_{vv} W_{vh}) W_{hc},
    \label{eq_vgcn_final}
\end{equation}

\noindent where $m$ is the mini-batch size, $v$ is the vocabulary size, $h$ is the hidden layer size, $c$ the class size or sentence embedding size. Every row of $X_{mv}$ is a vector containing document features, which can be a bag-of-words vector, or word embedding of BERT. The above equation aims to produce a layer of convolution of the graph, which captures the part of the graph relevant to the input (through $X_{mv} \tilde{A}_{vv}$), then performs 2 layers of convolution, %The resulting GCN representation thus 
combining  words from input sentence with their related words in vocabulary graph.  

% Usually, for a given graph such as citation relationship, people use two layers of GCN/GNN to capture information about direct and indirect neighbors~\cite{kipf2017semi,howpowergnn}, but in our case, one layer is better than two layers, and our experiment confirms this. We believe that this is because the meaning of non-direct neighbors is too far from the original word. 

\subsection{Integrating VGCN into BERT}

% BERT~\cite{devlin2018BERT} is a excellent self-attention model. For the BERT model used for the text classification task, it can be divided into three parts. The first part is the word embedding module with the position information of the word, and the second part is the transformer module using multi-layer multi-head self-attention stacking, and the third part is the normal fully connected layer using the last sentence embedding for classification.
When BERT is applied to text classification, a typical solution contains three parts. The first part is the word embedding module with the position information of the word; the second part is the transformer module using multi-layer multi-head self-attention stacking; and the third part is the fully connected layer using the output sentence embedding for classification.

Self-attention operates with a query Q against a key K and value V pair. The attention score is calculated as follows:
\begin{equation}
    \text{Attention}(Q,K,V)=\text{Softmax}\left(\frac{QK^T}{\sqrt{d_k}}\right)V,
    \label{attentionEq}
\end{equation}
\noindent where the denominator is a scaling factor used to control the scale of the attention score, $d_k$ is the dimension of the query and key vectors. Using these attention scores, every word can get a weighted vector representation encoding the contextual information. 

Instead of using only word embeddings of the input sentence in BERT, we feed both the vocabulary graph embedding obtained by Equation~\ref{eq_vgcn_final} and the sequence of word embeddings to BERT transformer. This way, not only the order of the words in the sentence is retained, but also the background information obtained by VGCN is utilized. The overall VGCN-BERT model is schematically illustrated in Figure \ref{fig:framework}.
Through the attention score calculated by Equation~\ref{attentionEq}, local embedding and global embedding are fully integrated after layer-by-layer interaction in 12-layer and 12-heads self-attention encoder. The corresponding VGCN can then be formulated as:

\begin{equation}
    \textbf{G}_\textbf{embedding} = \text{ ReLU}( X_{mev} \tilde{A}_{vv} W_{vh}) W_{hg},
    \label{eq_vgcn_graph_embedding}
\end{equation}
\noindent where $W_{hg}$, which was originally used for classification, becomes the output of size $g$ of graph embedding (hyperparameter) whose dimension is the same as every word embedding; 
%It is equivalent to how many graph embedding we want to enrich for the sentence. 
$m$ is the size of the mini-batch; $e$ is the dimension of  word embedding, and $v$ is the vocabulary size.

% The overall VGCN-BERT model is schematically illustrated in Figure \ref{fig:framework}.

%Then we feed all the embedding vectors to the 12-layer and 12-heads self-attention encoder, in which each word in the sentence will pay different attention to other words as well as the vocabulary graph embedding, and achieve full interaction. Finally, we use the sentence embedding produced at 12th layer encoder for classification.

% In general, this method of flexible use of global graph information by self-attention mechanism is more powerful than a vanilla combination method, which simply concatenate the sentence embedding from BERT and the global embedding from GCN, and then outputting the classified score through a fully connected layer. We will compare with the vanilla combination model from~\cite{vanillaBertGCN} in the experimental part.

%%%%%%%%%%%%%%%%%%%%%%%%%%%%%%%%%
%%         Experiment          %%
%%%%%%%%%%%%%%%%%%%%%%%%%%%%%%%%%

\section{Experiment}
We evaluate VGCN-BERT and compare it with baseline models on 5 datasets to verify whether our model can leverage both local and global information. 
%on bringing useful global information to the model.

\subsection{Baselines}
In addition to the original BERT model, we also use several other neural network models as baselines.

\begin{itemize}
\item \textbf{MLP}: Multilayer perceptron with 2 hidden layers (512 and 100 nodes), and bag-of-words model with TF weighting.% or just term frequency. 
\item \textbf{Bi-LSTM}~\cite{biLSTMHinton}: The BERT's pre-trained word embeddings are used as input to the Bi-LSTM model.
\item \textbf{Text GCN}: The original Text GCN model uses the same input feature as MLP model, and we use  the same training parameters as in~\cite{yao2019graph}.
\item \textbf{VGCN}: This model only uses VGCN, corresponding to Equation \ref{eq_vgcn_graph_embedding}, but the output dimension becomes the class size. %to build a pure VGCN model. Similarly Bi-LSTM model, 
BERT's pre-trained word embeddings are used as input. The output of VGCN is relayed to a fully connected layer with Softmax function to produce the classification score. This model only uses the global information from vocabulary graph.
%can be thought of as a degenerated version of model VGCN-BERT, with BERT removed.
\item \textbf{BERT}: We use the small version (Bert-base-uncased) pre-trained BERT~\cite{devlin2018BERT}.
\item \textbf{Vanilla-VGCN-BERT}: Vanilla combination of BERT and VGCN is similar to  \cite{vanillaBertGCN}, which produces two separate representations through BERT and GCN, and then concatenates them.  %for the purpose of citation recommendation, the original model uses the paper citation relationship graph given by dataset, and we replace this graph with our vocabulary graph. The method is concatenate the last [CLS] embedding (ie BOS embedding) from the Self-Attention of BERT with the output embedding from VGCN, and then use 
ReLU and a fully connected layer are applied to the combined representation for classification. The main difference of this model with ours is that it does not allow  interactions between the input text and the graph.
\end{itemize}

\subsection{Datasets}
We ran our experiments on the following five datasets:
%across three different types of text classification tasks, i.e., sentiment classification, grammaticality detection, and hate speech detection. %All the texts we consider are short. This is mainly due to the limitation of our computational capacity. %It should be noted that due to the memory of the GPU graphics card, we cannot select a long text classification dataset.

\begin{itemize}
\item \textbf{SST-2} The Stanford Sentiment Treebank is a binary single-sentence classification task consisting of sentences extracted from movie reviews with human annotations of their sentiment~\cite{sstdataset}. We use the public version\footnote{https://github.com/kodenii/BERT-SST2} which contains 6,920 examples in training set, 872 in validation set and 1,821 in test set, for a total of 4,963  positive reviews and 4,650  negative reviews. The average length of reviews is 19.3 words.

\item \textbf{MR} is also a movie review dataset for binary sentiment classification, in which each review only contains one sentence ~\cite{pang2005seeing}\footnote{http://www.cs.cornell.edu/people/pabo/movie-review-data/}. We used the public version in~\cite{tang2015pte}\footnote{https://github.com/mnqu/PTE/tree/master/data/mr}. It contains 5,331 positive and 5,331 negative reviews. The average length is 21.0 words. 

\item \textbf{CoLA} The Corpus of Linguistic Acceptability is a binary single-sentence classification task. CoLA is manually annotated for acceptability (grammaticality)~\cite{warstadt2018neural}. We use the public version which contains  8,551 training data and 1,043 ation data\footnote{https://github.com/nyu-mll/GLUE-baselines}, for a total of 6,744  positive and 2,850  negative cases. The average length is 7.7 words. Since we do not have the label for the test set, we split 5\% of the training set as  validation set and use the original validation set as the test set.

\item \textbf{ArangoHate}~\cite{Arango2019HateNotEasy} is a resampled dataset merging the datasets from \cite{ZeerakWaseem2016dataset}  and \cite{Davidson2017dataset}. 
%The author of the new dataset intended to alleviate the user-overfitting issue. 
It contains 2,920 hateful documents and 4,086 normal documents. The average length is 13.3 words. Since the dataset is not pre-divided into training, validation and test sets, we randomly split it into three sets at the ratio of 85:5:10.

\item \textbf{FountaHate} is a large four-label dataset for hate speech and offensive language detection~\cite{founta2018largeDataset}. 
%The author constructed it intended to better approximate a real-world setting where abuse is relatively rare. 
It contains 99,996\footnote{The final version provided by the author is more than the one described in the paper. } tweets with cross-validated labels and is classified into 4 labels: normal (53,851), spam (14,030), hateful (27,150) and abusive (4,965). The average length is 15.7 words. %The dataset is obtained from its constructor.
%Because a lot of tweets have disappeared, there will be a big bias in using tweet id for crawling, the author kindly provides us the full version of the dataset. 
Since the dataset is not pre-divided into training, validation and test sets, we split it into three sets at the ratio of 85:5:10 after shuffle. 
\end{itemize}

% As the label distribution in datasets CoLA (2.37:1), ArangoHate (1.40:1) and FountaHate (10.85:5.47:2.83:1) are uneven,  we train all models using the weight of each class on these three datasets.

% Therefore we use the weighted cross entropy as loss function. The weight of each of the classes ($W_c$) is calculated by
% \begin{equation}
%     W_{c}=\frac{\#dataset}{\#classes\cdot\#one\_class},
% \end{equation}
% where $\#dataset$ is the total size of dataset and $\#classes$ is the number of classes and $\#one\_class$ is the count of one class.

\subsection{Preprocessing and setting}
% When training and testing different models, we mainly follow two principles: 1. Training with parameters that are most beneficial to a particular model (such as learning rate), 2. Training different models under the same conditions (such as the same Loss function, The same pre-processed data, the same random seed, the same L2 weight decay)

We removed URL strings and @-mentions to retain the text content, then the text was lower-cased and tokenized using NLTK's \textit{TweetTokenizer} \footnote{http://www.nltk.org/api/nltk.tokenize.html}. We use BERTTokenizer function to split text, so that the vocabulary for GCN is always a subset of pre-trained BERT's vocabulary. %We keep all words of all datasets for all models, except when evaluating MLP with FountaHate dataset, we remove the stop words defined in NLTK\footnote{http://www.nltk.org/}.
When computing NPMI on a dataset, the whole sentence is used as the text window to build the vocabulary graph. 
%for MR, CoLA and SST-2 datasets, and set window size as 20 for two hate speech datasets
The threshold of NPMI is set as 0.2 for all datasets to filter out non-meaningful relationships between words. 

% For the phase of pre-train the VGCN module, we set the learning rate as 4e-4 for CoLA, and 2e-4 for other datasets, we set $L_2$ loss weight decay as 1e-4 and total epoch as 9 for all datasets. We fixed the values of word embedding comes from BERT for pre-train the weights of the VGCN module.

In the VGCN-BERT model, the graph embedding output size is set as 16, and the hidden dimension of graph embedding as 128. We use the \textit{Bert-base-uncased} version of pre-trained BERT, and set the max sequence length as 200. The model is then trained in 9 epochs with a dropout rate of 0.2. The following are other parameter settings for different datasets.

%\begin{table*}[t]\footnotesize
%    \centering
%    \renewcommand{\arraystretch}{1.2}
%    \caption{Parameter setting.}
 %   \begin{tabular}{p{3cm}p{3cm}p{3cm}p{3cm}}
%    \toprule
%    \textbf{Collection} & \textbf{mini-batch} & \textbf{Learning rate} & \textbf{L2 weight decay}  \\ \hline
%    SST-2       & 16  & l2-5 & 0.01 \\
%    CoLA and MR & 16 & 8e-6 & 0.01 \\
 %   ArangoHate & 16 & 1e-5 & 1e-3 \\
%    FountaHate & 12 & 4e-6 & 2e-4\\
%\end{tabular}
%\end{table*}

\begin{itemize}
    \item  SST-2:  mini-batch = 16, learning rate = 1e-5,  $L_2$ loss weight decay = 0.01. 
    \item  CoLA and MR: mini-batch  = 16, learn. rate = 8e-6,  $L_2$ loss  decay = 0.01. 
    \item  ArangoHate: mini-batch  = 16, learn. rate = 1e-5, and $L_2$ loss  decay = 1e-3. 
    \item  FountaHate: mini-batch  = 12, learn. rate = 4e-6, and $L_2$ loss  decay = 2e-4. 
\end{itemize}

These parameters are set based on our preliminary tests. We also use the default fine-tuning learning rate and $L_2$ loss weight decay as in~\cite{devlin2018BERT}. % It should be noted that these may not be optimal for all models and all datasets.
The baseline methods are set with the same parameters as in the original papers.

%For the baseline of the original BERT model and  VGCN model, the parameter settings are the same as the VGCN-BERT.
%, except that there is no graph embedding output dimension. 
%For  MLP, we use the term-frequency (TF) as inputs feature, and set the dimensions of two layers of fully connected networks as 512 and 100, learning rate as 1.5e-3 and $L_2$ loss weight decay as 2e-5, batch size as 64, total train epoch as 100, early stopping as 10. 
%For the Text GCN, we use the same input feature as MLP, and  the same training parameters as~\cite{devlin2018BERT}. 
%200 total train epochs, 20 for early stopping, 200 hidden dimensions, 0.02 learning rate, and 0 for $L_2$ loss weight decay. 
%The only difference is that we set the dropout rate as 0.2 in order to be consistent with other models.
%For Bi-LSTM, we use the word embedding from pre-trained \textit{Bert-base-uncased} version , with the word embedding dimension 768, and we set hidden dimension of Bi-LSTM as 100, the fully connected layer dimensions as 50, learning rate as 1e-4 and $L_2$ loss weight decay as 0, batch size as 32, total train epochs as 20.

\subsection{Loss Function}
We use the cross-entropy as the loss function for all models, except for FountaHate dataset where we use the mean squared error as the loss function in order to leverage the annotators' voting information. 
% \zb{More detail, the ``vote'' column of FountaHate datset refers to the number of the annotators who have provided the majority label, then we divide it by the number of annotators to get the true ground confidence, and then we use MSE to fit this confidence.}

We use Adam as training optimizer for all models. For cases where the label distributions  are uneven
(CoLA (2.4:1), ArangoHate (1.4:1) and FountaHate (10.9:5.5:2.8:1)), \textit{comput\_class\_weight} function\footnote{https://scikit-learn.org/stable/modules/generated/sklearn.utils.class\_\\weight.compute\_class\_weight.html} from scikit-learn~\cite{sklearn_api} is used as the weighted loss function. The weight of each the classes ($W_c$) is calculated by
\begin{equation}
    W_{classes}=\frac{\#dataset}{\#classes\cdot\#every\_class},
\end{equation}
\noindent where $\#dataset$ is the total size of dataset and $\#classes$ is the number of classes and $\#every\_class$ is the count of every class. 

\subsection{Evaluation Metrics}
We adopt the two most widely used  metrics to evaluate the performance of the classifiers - the weighted average F1-score, and the macro F1-score~\cite{classificationevaluation}. 
%When training the model, we calculate the weighted average F1-score of the test set while the F1-score for validation set is the best. The weighted average F1-score and macro F1-score are according to the following formulas,
\begin{equation}
%\begin{split}
    \textbf{Weighted avg F1}=\sum_{i=1}^{C}{F1_{c_{i}}*W_{c_{i}}}, \hspace{12pt}
    \textbf{Macro F1}=\frac{1}{C}\sum_{i=1}^{C}{F1_{c_{i}}}
%\end{split}
\end{equation}

\subsection{Experimental Result}

The main results on weighted average F1-Score and macro F1-Score on test sets are presented in Table~\ref{tab:allResult}. The main observation is that VGCN-BERT outperforms all the baseline models (except against Vanilla-VGCN-BERT on MR dataset). In particular, it outperforms both VGCN and BERT alone, confirming the advantage to combine them.

Among the models that only use local information, we see that BERT outperforms MLP, Bi-LSTM. Between the models that exploit a vocabulary graph - VGCN and Text-GCN, the performance is similar.

Vanilla-VGCN-BERT and VGCN-BERT are two models that combine local and global information. In general, these models perform better than the other baseline models. This result confirms the benefit of combining local information and global information.

Comparing VGCN-BERT with Vanilla-VGCN-BERT, we see that the former generally performs better. The difference is due to the interactions between local and global information. The superior performance of VGCN-BERT clearly shows the benefit of allowing interactions between the two types of information.

%It is also interesting to note that the results we obtained with VGCN-BERT on two datasets are superior to the best performance achieved in previous studies: on ArangoHate - the previous highest avg. F1-score of 79.64 in \cite{Arango2019HateNotEasy} and on FountaHate - the previous highest weighted avg. F1-score of 80.5 in \cite{lee2018comparative}. This shows that our model is competitive against the state of the art.

%see VGCN model using only vocablulary graph outperform MLP model and Bi-LSTM model on four datasets of SST-2, MR, CoLA and ArangoHate.  
%The Vanilla-VGCN-BERT don't always outperform BERT, but the VGCN-BERT outperform BERT on all datasets, and outperform Vanilla-VGCN-BERT on four datasets of SST-2, CoLA, ArangoHate and FountaHate.

% It should be noted that we can't ensure that such parameter settings are optimal for all models and all datasets. These datasets have a large difference in quantity and label distribution, especially the first three data sets are relatively small, and a small change in the parameters of learning rate and $L2$ decay will result in large fluctuations in performance.  Despite this, our performances on datasets of ArangoHate and FountaHate reach the state-of-the-art~\footnote{To our best knowledge, the highest weighted avg. F1-score on test set of ArangoHate dataset is 79.64 in \cite{Arango2019HateNotEasy}, and the highest weighted avg. }}.

\begin{table*}[t]\footnotesize
    \centering
    \renewcommand{\arraystretch}{1.2}
    \caption{Weighted average F1-Score and (Macro F1-score) on the test sets. We run 5 times under the same preprocessing and random seed. Macro F1-score and Weighted F1-Score are the same on SST-2 and MR. 
    %, because the labels of test set are even. 
    Bold indicates the highest score and underline indicates the second highest score.
    %\zb{The last three lines show the state-of-the-art claimed by other papers in 2019. Note that the number of VLAWE and XLNET are the accuracy metric. We reach the state-of-the-art on \textit{FountaHate}.} 
    }
    \begin{tabular}{p{3.2cm}p{1.1cm}p{1cm}p{2.1cm}p{2.1cm}p{2.1cm}}
    \toprule
    \textbf{Model} & \textbf{SST-2} & \textbf{MR} & \textbf{CoLA} & \textbf{ArangoHate} & \textbf{FountaHate} \\ \hline
    MLP       & 80.78  & 75.55 & 61.39 (53.20) & 84.71 (84.42) & 79.22 (65.33) \\
    Text-GCN   & 80.45  & 75.67 & 56.18 (52.30) & 84.77 (84.43) & 78.74 (64.54) \\
    Bi-LSTM   & 81.32  & 76.39 & 62.88 (55.25) & 84.92 (84.58) & 79.04 (65.13) \\
    VGCN & 81.64 & 76.42 & 63.59 (54.82) & 85.97 (85.69) & 79.00 (64.04) \\
    BERT      & \underline{91.49}  & 86.24 & \underline{81.22} (\underline{77.02}) & 87.99 (87.75) & 80.59 (66.61) \\
    Vanilla-VGCN-BERT & 91.38 & \textbf{86.49} & 80.70 (76.30) & \underline{88.01} (\underline{87.79}) & \underline{81.11} (\underline{67.86}) \\
    VGCN-BERT & \textbf{91.93} & \underline{86.35}  & \textbf{83.68} (\textbf{80.46}) & \textbf{88.43} (\textbf{88.22})& \textbf{81.26} (\textbf{68.45}) \\
    % \hline
    % RNN-LTC~\cite{lee2018comparative} & - & - & - & - & 80.5x \\
    % VLAWE~\cite{DBLP:conf/naacl/IonescuB19} & - & 93.3x & - & - & - \\
    % XLNET~\cite{DBLP:journals/corr/abs-1906-08237} & 96.8x & - & - & - & - \\
    \bottomrule
    \end{tabular}
    \label{tab:allResult}
\end{table*}

\begin{figure*}[!t]
  \centering
  %\subfigure[The words in bracket are the first two related words with the largest values corresponding the formula \ref{eq_vgcn_graph_embedding} for each graph embedding.]{\includegraphics[width=\textwidth]{examples/mr_guid_7584_[CLS]_layer_5_good.pdf}}
  
  \subfigure
  %[The words in bracket are the first two related words with the largest values corresponding the formula \ref{eq_vgcn_graph_embedding} for each graph embedding. Class 1 is positive.]
  {\includegraphics[width=\textwidth]{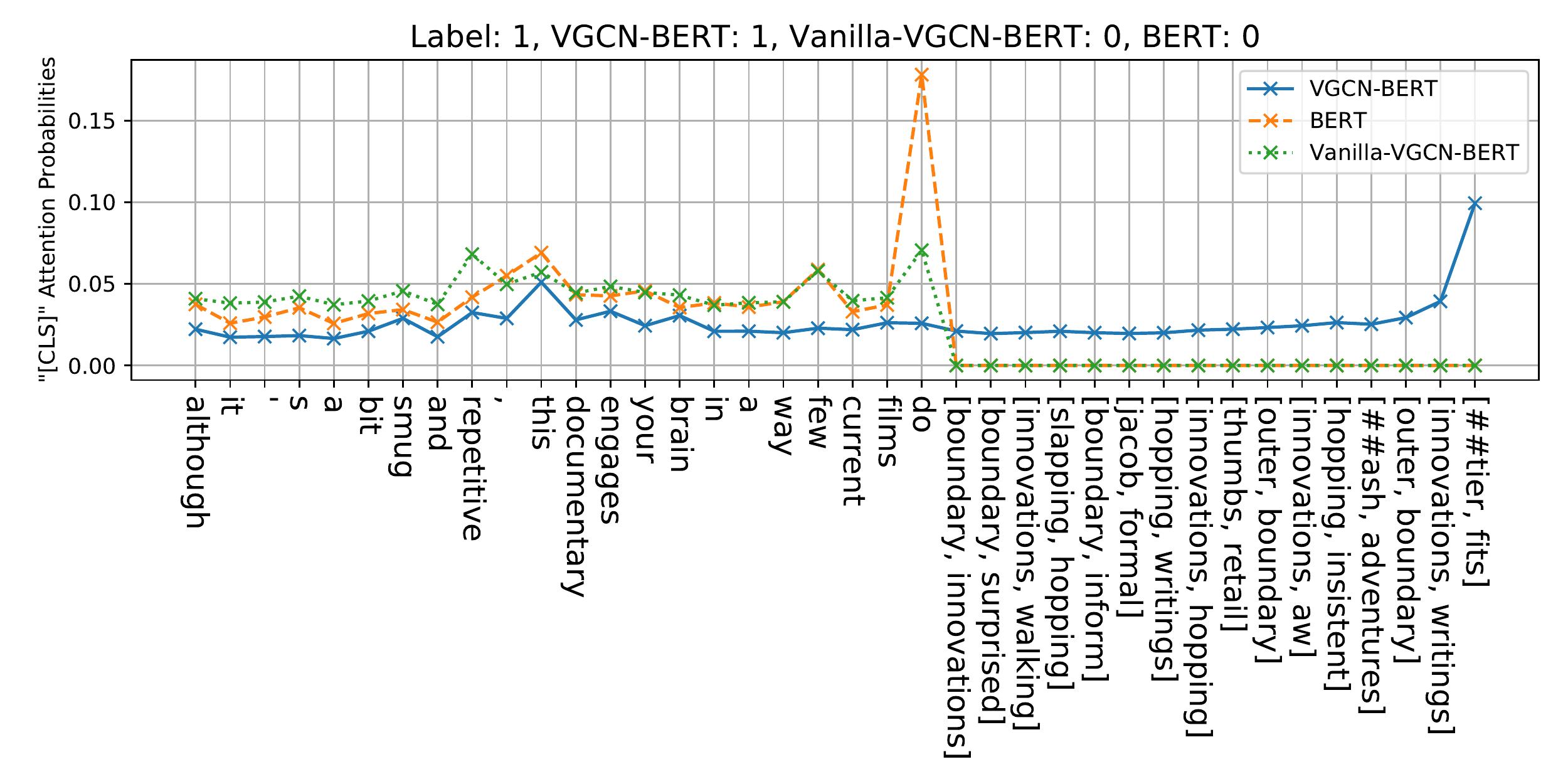}}
  
  \caption{Visualization of the attention that the token [CLS] (used as sentence embedding) pays to other tokens. %[CLS] is used as a sentence embedding for classification. %The y axis represents the attention probability,
  The first part corresponds to word embeddings of the sentence. The second part is the graph embedding. [word1, word2] indicates the approximate meaning of a dimension in graph embedding.
  %most important background word related to this sentence obtained from the vocabulary graph. BERT and Vanilla-VGCN-BERT do not attention for the graph embedding, so in the figure, the attention to [word] is equal to zero.
  }
  \label{fig:visulize_examples}
\end{figure*}

% \begin{figure}
%     \centering
%     \includegraphics{figure/mr_shocked_0_6_convolved.pdf}
%     \caption{Neighbors of ``piss'' in the vocabulary graph of FountaHate. The larger the font, the higher the NPMI.}
%     \label{fig:wordcloud_shocked}
% \end{figure}
  
\subsection{Visualization}
To better understand the behaviors of BERT, and its combination with VGCN, we visualize the attention distribution of the [CLS] token in the self-attention module of BERT, VGCN-BERT and Vanilla-VGCN-BERT models. As the vocabulary graph is embedded into vectors of 16 dimensions, it is not obvious to show what meaning corresponds to each dimension. To facilitate our understanding, we show the top two words from the sub-graph related to the input sentence, which are strongly connected to each of the 16 dimensions of graph embedding. More specifically, w each word embedding of a document is input to Equation \ref{eq_vgcn_graph_embedding}, we only need to broadcast the result of $XA$ and element-multiply it by $W$ to obtain the representation value of the words involved. The equation for obtain the involved words' id is as follow:
\begin{align}
    Z &= (\vec{x}A)^T\odot W, \\
    \text{IDs involved}&=\arg\text{sort}(Z[:,g]), 
\end{align}
\noindent where $x$ is a document in row vector,  $g \in [1,G]$, $G=16$ is the size of graph embedding. For example, the first dimension of graph embedding shown in Fig. \ref{fig:visulize_examples} corresponds roughly to the meaning of \textit{``[boundary, innovations]''}.

%We also find neighbor words that have a high co-occurrence weight (NPMI) with the sentence in the vocabulary graph for the VGCN-BERT model. 
% Our target is to look at the changes in the attention probability of these important neighbor words to other local tokens as well as [CLS].
% It should be noted that since there are 12 layers of self-attention encoder module, the interaction between tokens is very complicated, so we only examine the [CLS] attention probability to other tokens. Because [CLS] is the last sentence embedding for classification, we believe its attention probability for other tokens is more representative.
%Before using [CLS] as a sentence embedding for classification, [CLS] will fuse other tokens with different attention probability. For VGCN-BERT model, some words that are not found in the local sentence will be convolved by the vocabulary graph embedding module, then after passing through self-attention module, [CLS] of VGCN-BERT model will also fuse these neighbor words with certain attention probability. 
In Fig. \ref{fig:visulize_examples}, we show the attention paid to each word (embedding) and each dimension of graph embedding (second part). As BERT does not use graph embedding, %and Vanilla-GCN-BERT combines graph embedding after the self-attention encoder, 
the attention paid to graph embedding is 0. In VGCN-BERT, we see that graph embedding draws an important part of attention.

For the movie review \textit{``Although it's a bit smug and repetitive, this documentary engages your brain in a way few current films do.''}, the first half of the sentence is explicitly negative, while the remaining part expresses a positive attitude in an implicit way, which makes the sentence difficult to judge.
%tend to be judged as negative by most classifiers. 
For this example, BERT pays a very high attention to \textit{``do"}, and a quite high attention to  \textit{``this"}. These words do not bear much meaning in sentiments.
The final classification results by BERT is 0 (negative) while the true label is 1 (positive).

Vanilla-VGCN-BERT concatenates graph embedding with BERT without interaction between them. We can see that still no attention is paid to graph embedding, showing that such a simplistic combination cannot effectively leverage vocabulary information.

Finally, for VGCN-BERT, we see that a considerable part of attention is paid to graph embedding. The graph embedding is produced by integrating gradually the local information in the sentence with the global information in the graph. At the end, several dimensions of the graph embedding imply the meaning of \textit{``innovation"}, to which quite high attentions are paid. This results in classifying the sentence to the correct class (positive).

%With support of GCN component, we see [CLS] of BERT model and Vanilla-VGCN-BERT pay more attention to \textit{``repetitive''}, \textit{``this''}, \textit{``few''} and \textit{``do''}. Yet in the VGCN-BERT model, we see that the meaning of \textit{``[innovations]''} appears in several dimensions of graph embedding, which can be an explicit expansion of 
The meaning of \textit{``innovation"} is not produced immediately, but after a certain number of layers in BERT. In fact, through the layers of BERT, local information in the input sentence is combined to generate a higher level representation. In this example, at a certain layer, 
the expression \textit{``a way few current films do''} is grouped and represented as an embedding similar to the meaning of \textit{``innovation"}. From then, the meaning related to \textit{``innovation"} in the graph embedding is capture through self-attention, and reinforced later on through interactions between the local and global information. 

\section{Conclusion and Future Work}
In this study, we propose a new VGCN-BERT model to integrate a vocabulary graph embedding module with BERT. The goal is to complement the local information captured by BERT with the global information on the vocabulary, and allow both types of information to interact through the layers of attention mechanism. Our experiments on classification on 5 datasets show that the graph embedding does bring useful global information to BERT and this  improves the performance. In comparison with BERT and VGCN alone, our model can clearly lead to better results, showing that VGCN-BERT can indeed take advantage of both mechanisms.

As future work, we will consider using  other types of vocabulary graph such as Wordnet, in addition to a graph created by co-occurrences. We believe that Wordnet contains useful connections between words that NPMI cannot cover. It is thus possible to combine several lexical resources into the vocabulary graph.

\bibliographystyle{splncs04}
\bibliography{VGCN-BERT}

\end{document}